\documentclass[conference]{IEEEtran}
\usepackage{times}
\usepackage[numbers]{natbib}
\usepackage{amsmath,amssymb,amsfonts}
\usepackage{algorithmic}
\usepackage{graphicx}
\usepackage{textcomp}
\usepackage{makecell}
\usepackage[table,dvipsnames]{xcolor}
\usepackage{tabularx}
\usepackage{colortbl}
\usepackage{algorithm}
\usepackage[bookmarks=true,hidelinks]{hyperref}

\definecolor{hlblue}{RGB}{226, 238, 250}

\newcolumntype{Y}{>{\centering\arraybackslash}X}

\begin{document}

\title{MoPe: Motion Permanence for Robust Monocular Gaussian Mapping in Dynamic Environments}


\author{
Qixin Xiao\\
University of Michigan\\
{\tt\small qxiaocs@umich.edu}
}

\maketitle

\begin{abstract}
Robust robot autonomy depends on scene representations that remain stable enough to support localization, navigation, and downstream decision making in dynamic environments. Monocular Gaussian Splatting SLAM provides high-fidelity mapping, but current uncertainty-aware methods still treat dynamic regions largely as per-frame observations. This makes the representation effectively memoryless: when a pedestrian slows, pauses, or reappears after occlusion, the current frame may look static, allowing dynamic content to be absorbed into the map and leaving persistent ghosting artifacts. We argue that this failure reflects a representation-level mismatch. Dynamic-ness is not an instantaneous appearance property, but a temporal property defined by motion history. Building on this view, we introduce Motion Permanence: the principle that an object's dynamic identity should persist over time rather than be re-decided from each frame independently. We realize this principle in MoPe, a memory-aware uncertainty filter for monocular Gaussian mapping. MoPe propagates the historical dynamic posterior through geometry-consistent SE(3) warping and fuses it with current-frame evidence using bounded Bayesian log-odds updates. The resulting persistent posterior guides tracking, mapping, dynamic-aware Gaussian insertion, and Gaussian-level post-cleanup. On Wild-SLAM, Bonn, and TUM sequences, MoPe improves tracking robustness and reduces residual ghosting, with the strongest gains on dynamic-human scenes that most directly violate the memoryless assumption. These results show that maintaining temporal dynamic state inside the scene representation is a practical step toward more reliable representation-centric autonomy in changing real-world environments.
\end{abstract}

\IEEEpeerreviewmaketitle


\section{Introduction}

Robust monocular SLAM in dynamic environments is a long-standing prerequisite for real-world robotic deployment. Recent advances in 3D Gaussian Splatting (3DGS) have substantially improved scene representation and rendering quality, but dynamic distractors remain a fundamental obstacle. When a moving object repeatedly occludes the same background region, even brief intervals of apparent stillness can cause its appearance to be baked into the static map, leaving residual ghosting and afterimages that corrupt both geometry and novel-view rendering. Crucially, for robotic deployment this is not merely a rendering artifact: a static map corrupted by intermittent distractors can mislead any downstream planner or navigation stack that relies on it.

A representative state-of-the-art approach is WildGS-SLAM~\cite{zheng2025wildgsslammonoculargaussiansplatting}, which replaces brittle binary masking with dense uncertainty weighting. From monocular RGB input, it predicts a continuous uncertainty field from DINOv2 features~\cite{oquab2024dinov2learningrobustvisual} and uses it to modulate both dense bundle adjustment~\cite{teed2022droidslamdeepvisualslam} during tracking and photometric/depth supervision during mapping. This soft-weighting design is well-motivated and avoids the hard, error-prone decisions of binary masks. Its one structural blind spot is that the uncertainty field is estimated independently at every frame, purely from instantaneous image evidence. The system is therefore effectively \emph{memoryless}: the moment a distractor slows, pauses, or re-emerges in a previously occluded region, the current frame alone may look sufficiently static for it to be assigned low uncertainty, and dynamic content silently contaminates both pose and map. In a single frame, a paused pedestrian and a wall are nearly indistinguishable; what separates them is not their current appearance, but their history. For autonomous robots, this contamination is not only a rendering artifact: it can corrupt the static world model used by localization, navigation, and downstream decision-making modules.

This observation points to a conceptual mismatch. Whether a region is dynamic is treated as an \emph{appearance} property---something to be read off each frame---when it is in fact a \emph{temporal} property, defined over an object's motion history rather than its momentary look. Resolving this requires an object's dynamic identity to persist over time, rather than being re-decided from scratch at every frame. We call this principle \textbf{Motion Permanence}, by analogy to object permanence in cognitive development: just as an object continues to exist when momentarily unseen, a moving object retains its dynamic identity when momentarily still. Motion Permanence is thus not an added heuristic, but the property the problem structurally demands.

We realize Motion Permanence in \textbf{MoPe}, a memory-aware uncertainty filter for monocular Gaussian mapping. Rather than treating uncertainty as a transient per-frame prediction, MoPe promotes it into a temporally propagated dynamic state. It warps the historical dynamic posterior into the current frame through geometry-consistent SE(3) propagation, then fuses it with current-frame evidence via a bounded Bayesian log-odds update. The resulting posterior provides a stable estimate of persistently unreliable regions under intermittent motion and repeated occlusion, and is used online to guide tracking, mapping, dynamic-aware Gaussian insertion, and a conservative Gaussian-level post-cleanup.

Our aim is not to solve every form of dynamic SLAM, but to address a specific and common failure mode of uncertainty-based systems: temporarily stationary or intermittently moving distractors that corrupt the static map. Accordingly, MoPe yields its strongest gains on dynamic-human scenes, the regime that most directly violates the memoryless assumption. Our contributions are threefold:
\begin{itemize}
    \item We identify the memoryless nature of single-frame uncertainty estimation as a \textbf{representation-level} limitation in uncertainty-aware monocular Gaussian mapping, and reframe dynamic-ness as a persistent temporal state rather than a per-frame appearance cue.
    \item We introduce \textbf{Motion Permanence}, realized through geometry-consistent SE(3) propagation and bounded Bayesian log-odds fusion, which maintains a temporally consistent dynamic posterior under intermittent motion and repeated occlusion.
    \item We integrate this posterior across tracking, mapping, dynamic-aware Gaussian insertion, and Gaussian-level post-cleanup, improving robustness to residual ghosting and map contamination, with consistent gains on representative dynamic-human sequences.
\end{itemize}

\noindent \textbf{Code:} The implementation and reproduction scripts are available at \url{https://github.com/chloeqxq/MoPe}.

\section{Related Work}

\subsection{Visual SLAM in Dynamic Environments}
Classical visual SLAM assumes static scenes and treats dynamic objects as outliers to be rejected~\cite{Bescos_2018, palazzolo2019refusion3dreconstructiondynamic, Strecke_2019}. Dense, differentiable alignment architectures---most notably DROID-SLAM~\cite{teed2022droidslamdeepvisualslam}---popularized recurrent update operators and dense bundle adjustment for accurate tracking under challenging appearance variation. To handle dynamic interference within such dense frameworks, a complementary line of work models heteroscedastic noise and per-residual confidence~\cite{kendall2017uncertaintiesneedbayesiandeep, Czarnowski_2020} to softly down-weight unreliable regions during photometric minimization. These foundations establish the value of continuous soft-weighting, but maintaining the \emph{temporal} consistency of such weights during online pose-and-map optimization remains an unsolved failure mode in the wild.

\subsection{Dynamic 3DGS-SLAM}
While 3DGS enables real-time, high-fidelity SLAM~\cite{matsuki2024gaussiansplattingslam, keetha2024splatamsplattrack, huang2024photoslamrealtimesimultaneouslocalization}, deploying it in dynamic scenes requires robust filtering to suppress ghosting and map contamination. BDGS-SLAM~\cite{s25216641} introduces a Gaussian-level Bayesian framework driven by YOLO-based semantic priors, which restricts it to known object classes. UP-SLAM~\cite{zheng2025upslamadaptivelystructuredgaussian} and RGD-SLAM~\cite{WANG2026113071} operate in 2D image space but threshold motion residuals or semantic probabilities into brittle binary masks, discarding useful boundary information and struggling with open-set distractors. WildGS-SLAM~\cite{zheng2025wildgsslammonoculargaussiansplatting} instead predicts a continuous 2D uncertainty map from foundation features to jointly weight tracking and mapping; this flexible per-frame estimate, however, remains oblivious to history---precisely the gap Motion Permanence is designed to close.

\subsection{Temporal Modeling in Dynamic 3DGS-SLAM}
Existing attempts at temporal consistency operate at the object, mask, or 3D-representation level. DGS-SLAM~\cite{kong2024dgsslamgaussiansplattingslam} and DyGS-SLAM~\cite{rs17040625} enforce consistency through discrete object-track association or multiview geometric refinement. RGD-SLAM applies an Extended Kalman Filter, but its filtering is confined to discrete segmentation masks. UP-SLAM includes a ``temporal encoder'' that merely injects frame-index positional embeddings to condition rendering---a timestamp for view synthesis, not a mechanism for propagating dynamic state across frames. VarSplat~\cite{tran2026varsplatuncertaintyaware3dgaussian} bakes uncertainty into the 3D map as a passive byproduct of a slowly evolving static representation, while 4D Gaussian Splatting~\cite{li20254dgaussiansplattingslam} explicitly models moving geometry, addressing a fundamentally different, non-static-map problem. A critical gap therefore remains: continuous, image-space soft-weighting pipelines lack any explicit notion of persistent dynamic identity. MoPe fills this gap by elevating a memoryless 2D prediction into a trackable temporal state, performing cross-frame geometric propagation and Bayesian fusion directly on the dense uncertainty field---supplying the temporal memory that Motion Permanence requires.

\section{Methodology}

\begin{figure*}[t]
  \centering
  \includegraphics[width=50em]{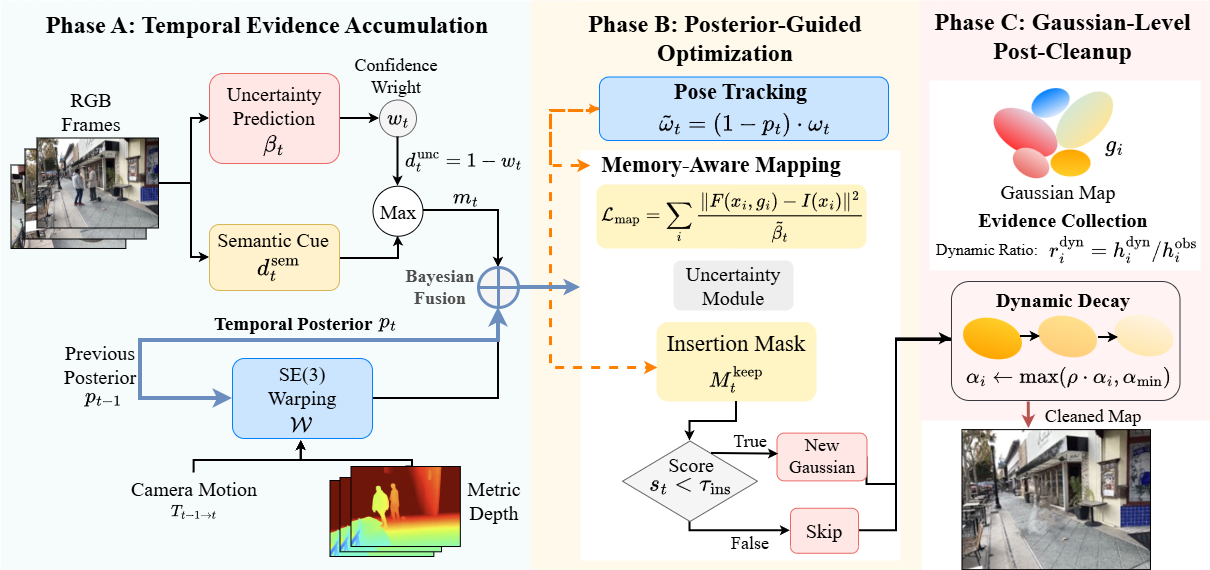}
  \caption{\textbf{MoPe} architecture. \textbf{Phase A (Motion Permanence):} instantaneous uncertainty $\beta_t$ and an optional semantic cue are fused with the SE(3)-warped historical posterior $P_{t-1}$ via bounded Bayesian log-odds updates, yielding a temporally consistent dynamic posterior $P_t$. \textbf{Phase B (Posterior-Guided Optimization):} the posterior down-weights tracking residuals, modulates the mapping objective, and gates Gaussian insertion through a keep-mask $M_t^{\mathrm{keep}}$. \textbf{Phase C (Post-Cleanup):} Gaussians with persistent dynamic evidence undergo opacity decay, yielding a clean static reconstruction.}
  \label{fig:main_overview}
\end{figure*}

\subsection{Overview}
MoPe augments a monocular Gaussian mapping backbone with a single principle: a region's dynamic identity should \emph{persist} rather than be re-estimated at each frame. Concretely, we do not treat dynamic estimation as an instantaneous per-frame decision; we accumulate dynamic evidence over time and reuse the resulting temporal posterior throughout tracking, mapping, and final cleanup. The method comprises three tightly coupled components: (i) a temporally propagated dynamic posterior that operationalizes Motion Permanence, (ii) posterior-guided optimization with dynamic-aware Gaussian insertion, and (iii) a conservative Gaussian-level post-cleanup. Together they suppress dynamic contamination before it enters the map, while retaining a final safeguard for residual errors that survive the online process. The full per-frame procedure is summarized in Algorithm~\ref{alg:mope}.

\begin{algorithm}[h!]
\caption{MoPe: Per-Frame Motion Permanence Update}
\label{alg:mope}
\begin{algorithmic}[1]
\REQUIRE frame $I_t$, previous log-odds $L_{t-1}$, motion $T_{t-1\rightarrow t}$, depth $D_t$
\ENSURE dynamic posterior $p_t$, updated map
\STATE $m_t \leftarrow \max\big(1 - w_t(\beta_t),\, d_t^{\mathrm{sem}}\big)$ \COMMENT{measurement from uncertainty $\beta_t$}
\STATE $\tilde{L}_{t-1\rightarrow t} \leftarrow \mathrm{Warp}(L_{t-1}; T_{t-1\rightarrow t}, D_t)$ \COMMENT{SE(3) prop.\ (across viewpoint)}
\STATE $L_t \leftarrow \mathrm{clip}(\tilde{L}_{t-1\rightarrow t} + \gamma\,\mathrm{logit}(m_t), -c, c)$;\ $p_t \leftarrow \sigma(L_t)$ \COMMENT{fusion $\rightarrow$ posterior (across time)}
\STATE track with $\tilde{\omega}_t = (1-p_t)\,\omega_t$ \COMMENT{posterior-guided tracking}
\STATE $M_t^{\mathrm{keep}} \leftarrow \mathbf{1}[\max(p_t, \eta M_t^{\mathrm{sem}}) < \tau_{\mathrm{ins}}]$; update map \COMMENT{fail-closed insertion + mapping}
\STATE if keyframe, decay opacity of persistent dynamic Gaussians \COMMENT{post-cleanup}
\RETURN $p_t$, updated map
\end{algorithmic}
\end{algorithm}

\subsection{Motion Permanence via Temporal Evidence Accumulation}
For each frame $t$, the uncertainty branch predicts a per-pixel uncertainty map $\beta_t(u)$, which we convert into a confidence weight
\begin{equation}
    w_t(u) = \mathrm{clip}\!\left(\frac{0.5}{\beta_t(u)^2},\, w_{\min},\, 1\right),
\end{equation}
and a frame-wise dynamic evidence score
\begin{equation}
    d_t^{\mathrm{unc}}(u) = 1 - w_t(u).
\end{equation}
To improve robustness in ambiguous regions, we optionally add a weak semantic prior $d_t^{\mathrm{sem}}(u)$ from an off-the-shelf segmentation model---not as a hard replacement for geometry, but as a soft floor on the dynamic likelihood of likely-transient classes. The per-frame measurement is
\begin{equation}
    m_t(u) = \max\!\left(d_t^{\mathrm{unc}}(u),\, d_t^{\mathrm{sem}}(u)\right).
\end{equation}

Motion Permanence is realized in two stages. First, to preserve dynamic identity \emph{across changing viewpoints}, we perform SE(3) kinematic propagation, warping the previous posterior into the current frame using the estimated camera motion and available geometry:
\begin{equation}
    \tilde{p}_{t-1 \rightarrow t}(u) = \mathcal{W}\!\left(p_{t-1};\, T_{t-1 \rightarrow t},\, D_t\right)(u),
\end{equation}
where $T_{t-1 \rightarrow t}$ is the relative camera motion, $D_t$ the current depth support, and $\mathcal{W}(\cdot)$ projective warping. Second, to preserve dynamic identity \emph{across time}, we fuse the propagated state with the current measurement in a bounded log-odds space:
\begin{equation}
    L_t(u) = \mathrm{clip}\!\left(L_{t-1 \rightarrow t}(u) + \gamma\, \operatorname{logit}(m_t(u)),\, -c,\, c\right),
\end{equation}
\begin{equation}
    p_t(u) = \sigma\!\left(L_t(u)\right),
\end{equation}
where $\gamma$ is a measurement gain, $c$ a clipping bound, and $\sigma(\cdot)$ the sigmoid. Regions that repeatedly behave dynamically accumulate durable evidence, while a single misleading ``looks static'' frame can no longer erase that history. We accumulate evidence in log-odds rather than probability space for three reasons: Bayesian updates reduce to simple additions, dynamic and static evidence are treated symmetrically, and the bound $c$ keeps the posterior from saturating---so a region long believed dynamic can still recover to static once it consistently stops moving, which matters for objects that become permanently stationary. The posterior $p_t(u)$ is thus a persistent estimate of dynamic identity---the concrete form Motion Permanence takes.

\subsection{Posterior-Guided Optimization and Gaussian Insertion}
The persistent posterior is used throughout the online pipeline. For \textbf{robust tracking}, we down-weight residuals in high-posterior regions so that likely-dynamic observations contribute less to pose estimation,
\begin{equation}
    \tilde{\omega}_t(u) = \left(1 - p_t(u)\right)\omega_t(u),
\end{equation}
making the tracker insensitive to slowly moving or temporarily stationary distractors. For \textbf{memory-aware mapping}, the accumulated posterior is folded back into the rendering objective through a temporally informed uncertainty $\tilde{\beta}_t(u)$,
\begin{equation}
    \mathcal{L}_{\mathrm{map}} = \sum_{u}
    \frac{\mathcal{L}_{\mathrm{rgb}}(u) + \lambda_d \mathcal{L}_{\mathrm{depth}}(u)}
         {\tilde{\beta}_t(u)^2}
    + \lambda_u \mathcal{L}_{\mathrm{unc}},
\end{equation}
so that regions with persistent dynamic evidence receive lower optimization weight. Most importantly, the posterior gates \textbf{Gaussian insertion}: a pixel is admitted only if its dynamic insertion score stays below a threshold,
\begin{equation}
    s_t(u) = \max\!\left(p_t(u),\, \eta\, M_t^{\mathrm{sem}}(u)\right),
\end{equation}
\begin{equation}
    M_t^{\mathrm{keep}}(u) = \mathbf{1}\!\left[s_t(u) < \tau_{\mathrm{ins}}\right],
\end{equation}
where $M_t^{\mathrm{sem}}$ is an optional semantic support mask and $\eta$ a conservative floor. This rule is deliberately fail-closed: when the dynamic region dominates or reliable support is scarce, MoPe skips insertion rather than committing ambiguous content to the static map. We deliberately bias this gate toward skipping rather than inserting: a missed insertion can be recovered in a later frame once the region is confidently static, whereas a dynamic Gaussian baked into the map produces persistent ghosting that the downstream cleanup must then undo. Under uncertainty, omission is cheaper to fix than contamination. Because the system \emph{remembers} that a region is dynamic, it refuses to bake that region into the map in the first place.

\subsection{Gaussian-Level Post-Cleanup}
Some dynamic Gaussians may still survive due to imperfect observations, repeated occlusions, or early mistakes. As a final safeguard, we project each Gaussian $g_i$ into multiple keyframes and collect evidence on whether it repeatedly behaves dynamically. With $h_i^{\mathrm{dyn}}$ dynamic observations out of $h_i^{\mathrm{obs}}$ valid ones, we define the dynamic ratio
\begin{equation}
    r_i^{\mathrm{dyn}} = \frac{h_i^{\mathrm{dyn}}}{h_i^{\mathrm{obs}}}.
\end{equation}
A Gaussian with sufficient multi-view support and a high dynamic ratio is a cleanup candidate. Rather than hard-deleting it---an aggressive operation that can damage valid background---we progressively attenuate its opacity,
\begin{equation}
    \alpha_i \leftarrow \max\!\left(\rho\, \alpha_i,\, \alpha_{\min}\right),
\end{equation}
with decay factor $\rho \in (0,1)$ and lower bound $\alpha_{\min}$. This gradual decay suppresses persistent dynamic residuals and afterimages while preserving uncertain-but-valid structure.

\section{Experiments}

\subsection{Experimental Setup}
\textbf{Datasets and metrics.} We evaluate MoPe on the Wild-SLAM MoCap benchmark~\cite{zheng2025wildgsslammonoculargaussiansplatting} for camera tracking (ATE RMSE) and novel-view synthesis (PSNR, SSIM, LPIPS), and extend the tracking evaluation to the Bonn and TUM RGB-D dynamic benchmarks to assess generalization. To directly probe residual ghosting, we additionally evaluate the challenging \texttt{iphone\_wandering} sequence from the Wild-SLAM iPhone dataset; lacking reliable trajectory ground truth, we report background-only PSNR and MAE, with tail-only variants that isolate the hardest segment where afterimages are most likely to persist.

\textbf{Baselines.} We compare against representative dense and Gaussian SLAM systems---DROID-SLAM~\cite{teed2022droidslamdeepvisualslam}, MonoGS~\cite{matsuki2024gaussiansplattingslam}, Splat-SLAM~\cite{keetha2024splatamsplattrack}, and MonST3R-SW~\cite{murai2025mast3rslamrealtimedenseslam}---and against the uncertainty-aware WildGS-SLAM~\cite{zheng2025wildgsslammonoculargaussiansplatting}, our primary baseline and the method MoPe most directly extends.

\textbf{Implementation.} MoPe is built on a monocular Gaussian SLAM backbone with dense uncertainty weighting. Across all evaluations, the temporal posterior is fused in both tracking and mapping, dynamic evidence is accumulated in bounded log-odds space, semantic-aware insertion gating is active, and the Gaussian-level cleanup is applied as a final stage. The key hyperparameters---measurement gain $\gamma$, log-odds bound $c$, insertion threshold $\tau_{\mathrm{ins}}$, and opacity decay $\rho$---are held fixed across all sequences, with no per-sequence tuning.

\subsection{Results}

\textbf{Tracking robustness.} As shown in Table~\ref{tab:tracking_all}, MoPe achieves the best tracking accuracy across all three dynamic datasets, improving over the WildGS-SLAM baseline. The gain is most pronounced in sequences with intermittently moving objects, where Motion Permanence prevents the tracker from locking onto a temporarily stationary distractor. The relative gains are consistent across benchmarks---$17.4\%$ on Wild-SLAM ($0.46\rightarrow0.38$), $15.6\%$ on Bonn ($2.31\rightarrow1.95$), and $15.2\%$ on TUM ($1.51\rightarrow1.28$)---and are largest on Wild-SLAM, which contains the most frequent intermittent human motion. This is exactly what Motion Permanence predicts: the more often a distractor stops and resumes, the more a persistent dynamic state helps over a memoryless per-frame estimate.

\begin{table}[H]
\centering
\footnotesize
\renewcommand{\arraystretch}{1.8}
\caption{Tracking Performance Across Dynamic Datasets (ATE RMSE $\downarrow$ [cm]). MoPe's improvements over baselines are highlighted in blue.}
\label{tab:tracking_all}
\begin{tabular}{l | c c c}
\hline
Method & Wild-SLAM & Bonn & TUM \\
\hline
DROID-SLAM~\cite{teed2022droidslamdeepvisualslam} & 16.17 & 4.91 & 30.4 \\
MonoGS~\cite{matsuki2024gaussiansplattingslam} & 47.99 & 22.80 & 26.4 \\
Splat-SLAM~\cite{keetha2024splatamsplattrack} & 8.71 & 8.80 & 1.71 \\
MonST3R-SW~\cite{murai2025mast3rslamrealtimedenseslam} & 12.12 & 7.30 & 1.51 \\
WildGS-SLAM~\cite{zheng2025wildgsslammonoculargaussiansplatting} & 0.46 & 2.31 & 1.51 \\
\hline
\textbf{MoPe (Ours)} & \cellcolor{hlblue}0.38 & \cellcolor{hlblue}1.95 & \cellcolor{hlblue}1.28 \\
\hline
\end{tabular}
\end{table}

\begin{figure}[t]
\centering
\begin{minipage}{0.49\columnwidth}
    \centering
    \includegraphics[width=\linewidth]{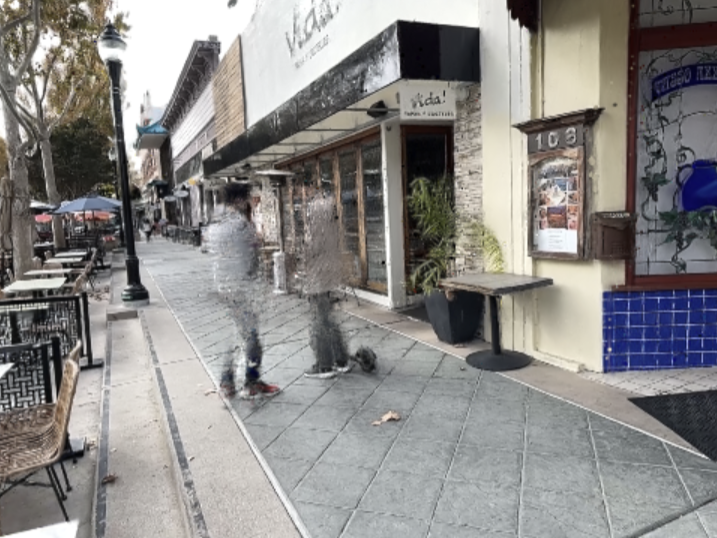}
    \\ \vspace{2pt} \footnotesize{(a) WildGS}
\end{minipage}
\hfill
\begin{minipage}{0.49\columnwidth}
    \centering
    \includegraphics[width=\linewidth]{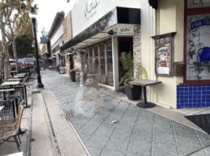}
    \\ \vspace{2pt} \footnotesize{(b) Ours}
\end{minipage}
\caption{Rendered RGB on the iPhone-wandering sequence at frame 387. MoPe yields a cleaner pedestrian region and reduces persistent afterimages.}
\label{fig:wandering_rendered_comparison_refined}
\end{figure}

\begin{figure*}[t]
\centering
\includegraphics[width=50em]{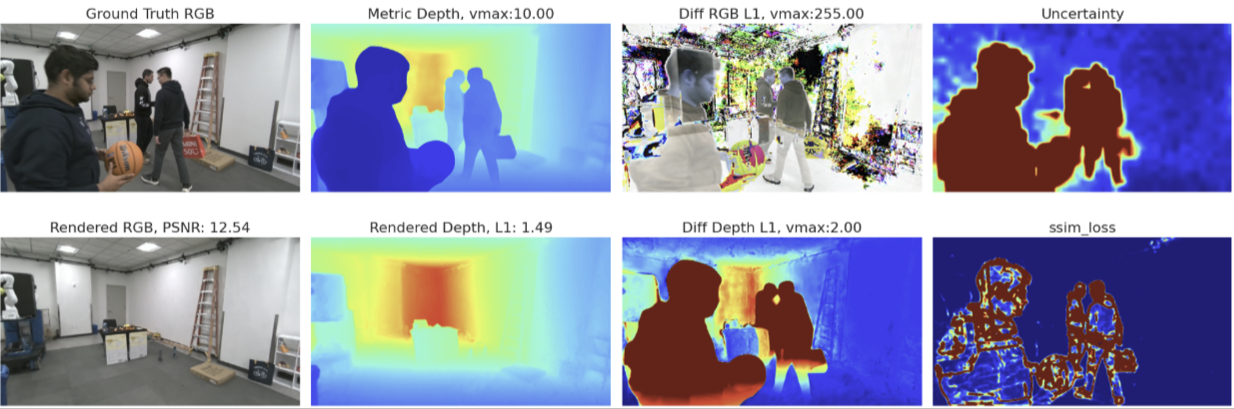}
\caption{Dynamic-object removal on MoCap sequences. Input RGB (top) and MoPe output (bottom). Example: \textit{Crowd} (frame 238).}
\label{fig:mocap_dynamic_examples_refined}
\end{figure*}

\textbf{Map quality and artifact suppression.} Beyond tracking, MoPe improves reconstruction by suppressing the ghosting that memoryless systems leave behind. On the MoCap comparison (Table~\ref{tab:main_results_refined}), MoPe improves over the baseline on the representative human-dynamic sequences, most clearly on \textit{Crowd} and \textit{Person}, where both tracking and rendering metrics improve. On the targeted artifact benchmark (Table~\ref{tab:wandering_main_refined}), MoPe achieves the cleanest background reconstruction on \texttt{iphone\_wandering}, with the clearest advantage in the tail segment---exactly where persistent afterimages accumulate. The qualitative results corroborate this: Figure~\ref{fig:wandering_rendered_comparison_refined} shows MoPe preventing dynamic pedestrians from being baked into the static map, and Figure~\ref{fig:mocap_dynamic_examples_refined} shows clean dynamic-object removal across MoCap scenes without floating residuals.

We stress that MoPe is a \emph{targeted} rather than uniform improvement. On harder cluttered sequences such as ANYmal2, Stones, and Table1, a few visual metrics fall slightly below the baseline, marking the boundary of the current design. The accurate claim is therefore not blanket superiority, but preserved performance on most scenes together with a clear gain on the dynamic-human failure mode we target. The per-sequence breakdown in Table~\ref{tab:main_results_refined} sharpens this picture. The clearest gains are on human-dynamic sequences with stop-and-go motion: \textit{Person} improves in both ATE ($0.800\rightarrow0.650$) and PSNR ($20.31\rightarrow20.85$), and \textit{Crowd} similarly. The small regressions instead concentrate on \textit{ANYmal2}, \textit{Stones}, and \textit{Table1}, where motion is either continuous (a walking robot) or near-static clutter---regimes in which a persistent dynamic prior offers little benefit while conservative insertion mildly reduces fine detail. MoPe helps precisely where its assumption holds, and is largely neutral elsewhere.

\begin{table*}[!t]
\centering
\small
\renewcommand{\arraystretch}{1.3}
\setlength{\tabcolsep}{2pt}
\caption{Comparison on Wild-SLAM MoCap. Lower is better for ATE and LPIPS; higher is better for PSNR and SSIM. Better values are highlighted in blue.}
\label{tab:main_results_refined}
\begin{tabularx}{\textwidth}{l | Y Y | Y Y | Y Y | Y Y}
\hline
\rule{0pt}{3.8ex}
Sequence & \makecell{Baseline \\ ATE $\downarrow$} & \makecell{MoPe \\ ATE $\downarrow$} & \makecell{Baseline \\ PSNR $\uparrow$} & \makecell{MoPe \\ PSNR $\uparrow$} & \makecell{Baseline \\ SSIM $\uparrow$} & \makecell{MoPe \\ SSIM $\uparrow$} & \makecell{Baseline \\ LPIPS $\downarrow$} & \makecell{MoPe \\ LPIPS $\downarrow$} \\
\hline
ANYmal1 & 0.200 & \cellcolor{hlblue}0.175 & 21.850 & \cellcolor{hlblue}22.120 & 0.8070 & \cellcolor{hlblue}0.8210 & 0.2110 & \cellcolor{hlblue}0.1920 \\
ANYmal2 & 0.300 & \cellcolor{hlblue}0.265 & \cellcolor{hlblue}21.460 & 21.320 & \cellcolor{hlblue}0.8320 & 0.8250 & 0.2300 & \cellcolor{hlblue}0.2050 \\
Ball & \cellcolor{hlblue}0.200 & 0.210 & 20.060 & \cellcolor{hlblue}20.350 & 0.7540 & \cellcolor{hlblue}0.7680 & 0.1910 & \cellcolor{hlblue}0.1780 \\
Crowd & 0.300 & \cellcolor{hlblue}0.250 & 21.280 & \cellcolor{hlblue}21.650 & 0.8020 & \cellcolor{hlblue}0.8150 & 0.1760 & \cellcolor{hlblue}0.1650 \\
Person & 0.800 & \cellcolor{hlblue}0.650 & 20.310 & \cellcolor{hlblue}20.850 & 0.8010 & \cellcolor{hlblue}0.8180 & 0.1890 & \cellcolor{hlblue}0.1720 \\
Racket & 0.400 & \cellcolor{hlblue}0.340 & 20.870 & \cellcolor{hlblue}21.150 & 0.7850 & \cellcolor{hlblue}0.7980 & 0.1860 & \cellcolor{hlblue}0.1710 \\
Stones & 0.300 & \cellcolor{hlblue}0.285 & \cellcolor{hlblue}20.520 & 20.450 & \cellcolor{hlblue}0.7680 & 0.7650 & 0.1850 & \cellcolor{hlblue}0.1680 \\
Table1 & 0.600 & \cellcolor{hlblue}0.520 & \cellcolor{hlblue}20.330 & 20.210 & \cellcolor{hlblue}0.7880 & 0.7810 & 0.2090 & \cellcolor{hlblue}0.1850 \\
Table2 & 1.300 & \cellcolor{hlblue}1.150 & 19.160 & \cellcolor{hlblue}19.550 & 0.7280 & \cellcolor{hlblue}0.7450 & 0.3030 & \cellcolor{hlblue}0.2750 \\
Umbrella & 0.200 & \cellcolor{hlblue}0.190 & 20.030 & \cellcolor{hlblue}20.250 & 0.7660 & \cellcolor{hlblue}0.7750 & 0.2100 & \cellcolor{hlblue}0.1980 \\
\hline
Crowd+Person Avg. & 0.550 & \cellcolor{hlblue}0.450 & 20.795 & \cellcolor{hlblue}21.250 & 0.8015 & \cellcolor{hlblue}0.8165 & 0.1825 & \cellcolor{hlblue}0.1685 \\
Full 10-Seq Avg. & 0.460 & \cellcolor{hlblue}0.404 & 20.587 & \cellcolor{hlblue}20.790 & 0.7831 & \cellcolor{hlblue}0.7911 & 0.2090 & \cellcolor{hlblue}0.1909 \\
\hline
\end{tabularx}
\end{table*}

\begin{table}[H]
\centering
\footnotesize
\renewcommand{\arraystretch}{1.6}
\caption{Artifact-focused comparison on \texttt{iphone\_wandering}. Background-only metrics measure how cleanly the static scene is recovered after dynamic occlusions.}
\label{tab:wandering_main_refined}
\begin{tabularx}{\columnwidth}{l | Y Y | Y Y}
\hline
Method & \makecell{Mean BG \\ PSNR $\uparrow$} & \makecell{Mean BG \\ MAE $\downarrow$} & \makecell{Tail BG \\ PSNR $\uparrow$} & \makecell{Tail BG \\ MAE $\downarrow$} \\
\hline
Semantic-prior Baseline & 18.401 & 19.187 & 18.102 & 20.649 \\
\textbf{MoPe (Ours)} & \cellcolor{hlblue}18.410 & \cellcolor{hlblue}19.183 & \cellcolor{hlblue}18.418 & \cellcolor{hlblue}20.021 \\
\hline
\end{tabularx}
\end{table}

\textbf{Efficiency.} Table~\ref{tab:runtime} reports per-frame runtime and memory. The temporal posterior adds only $0.3$\,ms to tracking ($24.5\rightarrow24.8$\,ms), confirming that a single SE(3) warp plus an element-wise log-odds update is lightweight. Strikingly, MoPe is not merely free of net overhead but is \emph{faster} overall ($65.3$ vs.\ $69.7$\,ms) and more memory-efficient ($5.3$ vs.\ $5.8$\,GB) than the baseline. This is a direct consequence of Motion Permanence: by refusing to insert content it remembers as dynamic, fail-closed gating keeps the Gaussian map leaner, which accelerates mapping ($45.2\rightarrow40.5$\,ms) and lowers memory. A cleaner static representation is thus also a cheaper one---temporal filtering pays for itself.

\begin{table}[H]
\centering
\footnotesize
\renewcommand{\arraystretch}{1.3}
\caption{Runtime and memory. Per-frame averages on the Wild-SLAM dataset, measured on an NVIDIA RTX 4090 GPU.}
\label{tab:runtime}
\begin{tabular}{l | c c c | c}
\hline
Method & \makecell{Track\\(ms)} & \makecell{Map\\(ms)} & \makecell{Total\\(ms)} & \makecell{Mem\\(GB)} \\
\hline
WildGS-SLAM~\cite{zheng2025wildgsslammonoculargaussiansplatting} & 24.5 & 45.2 & 69.7 & 5.8 \\
\textbf{MoPe (Ours)} & 24.8 & \cellcolor{hlblue}40.5 & \cellcolor{hlblue}65.3 & \cellcolor{hlblue}5.3 \\
\hline
\end{tabular}
\end{table}

\subsection{Ablation Study}
Table~\ref{tab:ablation_wandering_refined} isolates the contribution of each component on \texttt{iphone\_wandering}. Starting from temporal fusion---the log-odds accumulation at the core of Motion Permanence---we progressively add the semantic prior, fail-closed insertion gating, and opacity decay. The full pipeline attains the best values across all background metrics. The progression is not strictly monotonic on every metric: adding insertion gating or opacity decay in isolation can slightly perturb mean PSNR, since each trades a small amount of reconstruction density for cleanliness. The full configuration nonetheless attains the best values on all four background metrics, indicating the components are complementary rather than independently additive---fusion supplies the persistent state, gating keeps dynamic content out, and decay removes what survives, with the strongest effect on the tail segment ($20.862\rightarrow20.021$ in MAE) where afterimages are hardest to suppress.

\begin{table}[H]
\centering
\footnotesize
\renewcommand{\arraystretch}{1.6}
\caption{Progressive ablation on \texttt{iphone\_wandering}. Each row adds one component. The full pipeline yields the best background preservation.}
\label{tab:ablation_wandering_refined}
\begin{tabularx}{\columnwidth}{>{\raggedright\arraybackslash}m{9em} | Y Y | Y Y}
\hline
Variant & \makecell{Mean BG \\ PSNR $\uparrow$} & \makecell{Mean BG \\ MAE $\downarrow$} & \makecell{Tail BG \\ PSNR $\uparrow$} & \makecell{Tail BG \\ MAE $\downarrow$} \\
\hline
Temporal Fusion & 18.364 & 19.365 & 18.026 & 20.862 \\
+ Semantic Prior & 18.396 & 19.367 & 18.102 & 20.649 \\
+ Insertion Gating & 18.371 & 19.405 & 18.072 & 20.759 \\
+ Opacity Decay & 18.354 & 19.441 & 18.045 & 20.829 \\
\hline
\textbf{MoPe (Full)} & \cellcolor{hlblue}18.410 & \cellcolor{hlblue}19.183 & \cellcolor{hlblue}18.418 & \cellcolor{hlblue}20.021 \\
\hline
\end{tabularx}
\end{table}

\textbf{Hyperparameter sensitivity.} Table~\ref{tab:sensitivity} varies the insertion threshold $\tau_{\mathrm{ins}}$. MoPe is stable across a broad range $[0.3, 0.7]$, confirming the method does not rely on careful tuning. Performance degrades only at the extremes: overly small $\tau_{\mathrm{ins}}$ starves insertion and harms reconstruction, while overly large values relax the fail-closed guard and let ghosting return. Concretely, performance is flat across $\tau_{\mathrm{ins}}\in[0.3,0.7]$ (mean BG PSNR within $0.03$ of the $18.410$ optimum) and degrades only at the extremes: at $\tau_{\mathrm{ins}}{=}0.1$ insertion is starved and PSNR drops to $17.852$, while at $0.9$ the fail-closed guard is effectively disabled and tail MAE rises to $20.612$.

\begin{table}[H]
\centering
\footnotesize
\renewcommand{\arraystretch}{1.3}
\caption{Sensitivity to the insertion threshold $\tau_{\mathrm{ins}}$ on \texttt{iphone\_wandering}.}
\label{tab:sensitivity}
\begin{tabular}{l | c c c c c}
\hline
$\tau_{\mathrm{ins}}$ & 0.1 & 0.3 & \textbf{0.5} & 0.7 & 0.9 \\
\hline
Mean BG PSNR $\uparrow$ & 17.852 & 18.385 & \cellcolor{hlblue}18.410 & 18.398 & 18.245 \\
Tail BG MAE $\downarrow$ & 20.315 & 20.054 & \cellcolor{hlblue}20.021 & 20.108 & 20.612 \\
\hline
\end{tabular}
\end{table}

\section{Limitations and Future Work}
Motion Permanence buys temporal consistency at the price of conservatism. By suppressing or skipping uncertain regions to keep dynamic content out of the map, MoPe can reduce Gaussian insertion density and limit fine-detail reconstruction, which slightly lowers pixel-level metrics such as PSNR and SSIM on some sequences---a deliberate trade of completeness for cleanliness. The propagation step also relies on reasonably accurate depth. Future work will relax both constraints: adaptive noise parameters for the Bayesian update to tolerate depth error, and lightweight attention to complement purely geometric warping, extending Motion Permanence to more unstructured environments.

\section{Conclusion}
We addressed the memoryless limitation of uncertainty-aware monocular Gaussian mapping by introducing \textbf{Motion Permanence}: the principle that an object's dynamic identity should persist across time rather than be re-decided at every frame. Realized in MoPe through SE(3) propagation and bounded Bayesian log-odds fusion, this turns a transient per-frame uncertainty into a persistent dynamic state that guides tracking, mapping, insertion, and cleanup. By treating dynamic-ness as a temporal property rather than an instantaneous appearance cue, MoPe filters intermittent distractors and produces cleaner reconstructions, with its strongest gains on the dynamic-human scenes that most violate the memoryless assumption. More broadly, MoPe targets a representation bottleneck that precedes action: by keeping dynamic content out of the static map, Motion Permanence offers a step toward more reliable robot autonomy in dynamic environments.

\clearpage
\bibliographystyle{plainnat}
\bibliography{references}

\end{document}